\documentclass[10pt,twocolumn,letterpaper]{article}

\usepackage{cvpr}              % To produce the CAMERA-READY version

\usepackage{graphicx}
\usepackage{booktabs}
\usepackage{xcolor}      
\usepackage{colortbl}    
\usepackage{tabularx}
\usepackage{array}
\usepackage{xcolor}
\usepackage{multirow}
\usepackage{balance}
\usepackage{svg}
\svgpath{{figures/}}
\svgsetup{inkscapelatex=false,inkscapearea=drawing} % 

\definecolor{RowHL}{HTML}{EAF4FF}
\newcolumntype{Y}{>{\centering\arraybackslash}X}
\newcommand{\SAPC}{\mbox{SA,\,PC}}
\newcommand{\graycell}[1]{\textcolor{black}{#1}}

\newcommand{\grayrowA}[7]{\graycell{#1} & \graycell{#2} & \graycell{#3} & \graycell{#4} & \graycell{#5} & \graycell{#6} & \graycell{#7}}
\newcommand{\grayrowB}[6]{& \graycell{#1} & \graycell{#2} & \graycell{#3} & \graycell{#4} & \graycell{#5} & \graycell{#6} \\}

\definecolor{cvprblue}{rgb}{0.21,0.49,0.74}
\usepackage[pagebackref,breaklinks,colorlinks,allcolors=cvprblue]{hyperref}

\def\confName{CVPR}
\def\confYear{2026}

\title{Chain of Event-Centric Causal Thought for \\Physically Plausible Video Generation}

\author{
Zixuan Wang$^{1, \dagger}$ \quad
Yixin Hu$^{1, \dagger}$ \quad
Haolan Wang$^{1}$ \quad
Feng Chen$^{2, \dagger}$ \quad
Yan Liu$^{3}$ \quad \\
Wen Li$^{4}$ \quad
Yinjie Lei$^{1, *} $
\\
$^1$Sichuan University \quad
$^2$The University of Adelaide \\
$^3$Hong Kong Polytechnic University \quad
$^4$University of Electronic Science and Technology of China \\
{\tt\small zixuan98@stu.scu.edu.cn, yinjie@scu.edu.cn} \\
}
\begin{document}
\twocolumn[{%
\renewcommand\twocolumn[1][]{#1}%
\maketitle

\begin{center}
  \includegraphics[width=1.00\linewidth]{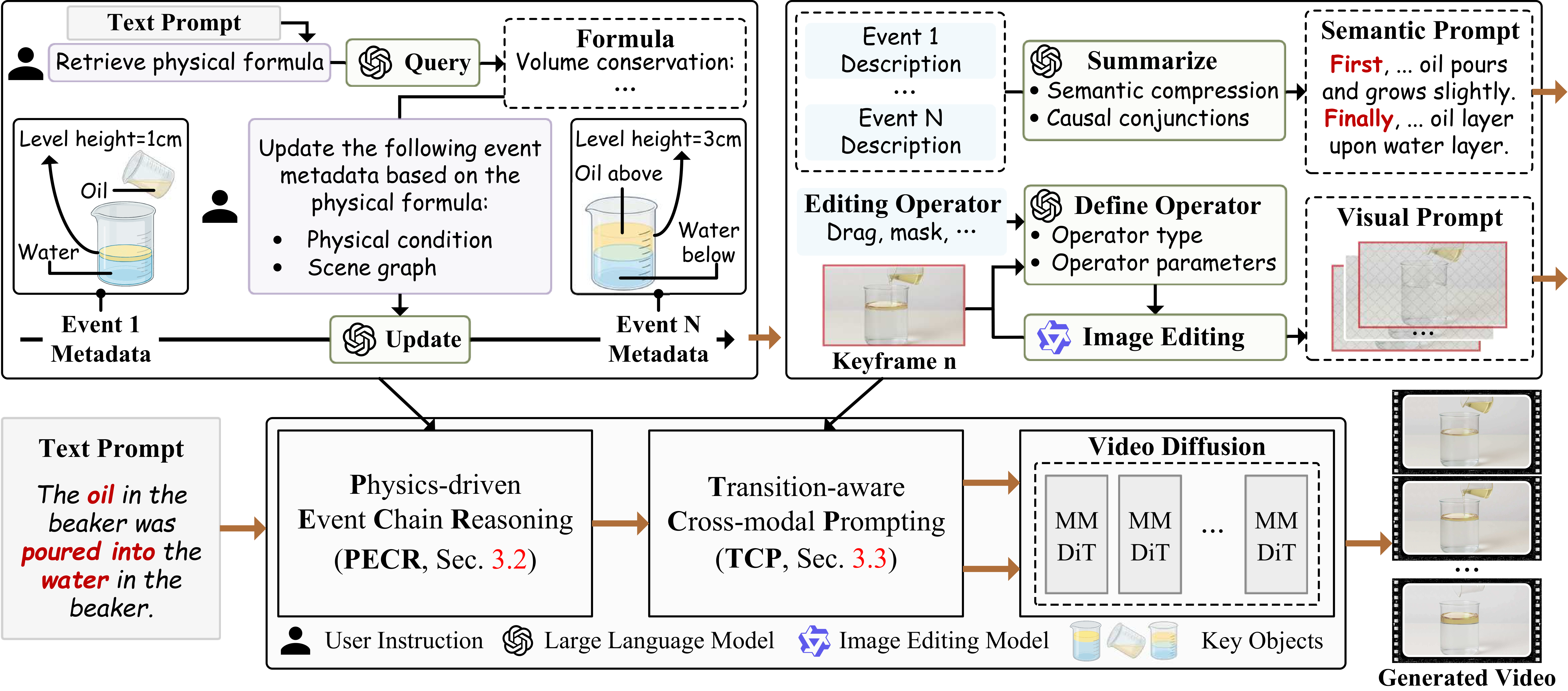}
  \captionof{figure}{Overview of our physically plausible video generation framework. We firstly decompose complex physical phenomena into a sequence of elementary events guided by physical formulas (~\cref{sec:3.1}), and secondly map logically ordered events to a holistic description and a set of keyframes, both of which are causally coherent (~\cref{sec:3.2}). Our inferred vision-language prompts enable off-the-shelf diffusion frameworks to generate videos capturing the causal progression of physical phenomena.
  }
  \label{fig:main_figure}
\end{center}
}]
\let\thefootnote\relax\footnotetext{\textsuperscript{$\dagger$} Equal contribution.}
\let\thefootnote\relax\footnotetext{\textsuperscript{*} Corresponding author.}
 \begin{abstract}
Physically Plausible Video Generation (PPVG) has emerged as a promising avenue for modeling real-world physical phenomena. PPVG requires an understanding of commonsense knowledge, which remains a challenge for video diffusion models. Current approaches leverage commonsense reasoning capability of large language models to embed physical concepts into prompts. However, generation models often render physical phenomena as a single moment defined by prompts, due to the lack of conditioning mechanisms for modeling causal progression. In this paper, we view PPVG as generating a sequence of causally connected and dynamically evolving events. To realize this paradigm, we design two key modules: (1) Physics-driven Event Chain Reasoning. This module decomposes the physical phenomena described in prompts into multiple elementary event units, leveraging chain-of-thought reasoning. To mitigate causal ambiguity, we embed physical formulas as constraints to impose deterministic causal dependencies during reasoning. (2) Transition-aware Cross-modal Prompting (TCP). To maintain continuity between events, this module transforms causal event units into temporally aligned vision-language prompts. It summarizes discrete event descriptions to obtain causally consistent narratives, while progressively synthesizing visual keyframes of individual events by interactive editing. Comprehensive experiments on PhyGenBench and VideoPhy benchmarks demonstrate that our framework achieves superior performance in generating physically plausible videos across diverse physical domains.  Code is available at \textcolor[RGB]{237,0, 140}{https://github.com/ZixuanWang0525/CoECT}.

\end{abstract}    
 \section{Introduction}
\label{sec:intro}

PPVG has opened up a wide range of real-world applications, including movie production \cite{zhang2025generative}, autonomous driving \cite{deng2024streetscapes}, and embodied AI \cite{agarwal2025cosmos}. In recent years, video diffusion models, such as Kling \cite{kling2024} and OpenAI-Sora \cite{liu2024sora}, have demonstrated remarkable capabilities in synthesizing photorealistic scenes from user prompts. However, brief prompts fail to provide the detailed physical laws required for the physically plausible generation. %Additionally, video diffusion models lack the ability to implicitly infer commonsense knowledge from these prompts. 
This hinders such generative models from simulating real-world physical phenomena, \textit{e.g.}, fluid dynamics, light refraction, and thermodynamic effects.  

Recent PPVG studies \cite{xue2025phyt2v,hao2025enhancing,zhang2025think} have augmented user prompts with physical concepts based on Large Language Model (LLM)-assisted reasoning. % The augmented prompts act as clearer physical constraints, guiding video generative models to render appearance and style with higher physical plausibility. 
However, these approaches typically simplify the generated physical phenomena to a single moment defined by static prompts. This challenge arises due to: (1) \textit{Causal Ambiguity}. In the real-world, physical phenomena unfold as causally ordered event units. Unfortunately, embedding a semantic tag to describe such complex phenomena often fails to capture their dynamic nature. This requires a structured decomposition of physical phenomena by causal deterministic reasoning. (2) \textit{Insufficient Physics-consistent Constraints.} Language alone is inherently incapable of conveying the causal continuous between events. Visual cues (\textit{e.g.}, reference videos) can provide observable evidence of event transitions. Even so, visual priors tightly aligned with specified physical phenomena are often hard to obtain.

In this paper, we propose an event-centric physically plausible video generation framework that models physical phenomena as transitions between causally linked events, as shown in ~\cref{fig:main_figure}. The framework consists of two core modules: (1) We design a Physics-driven Event Chain Reasoning (PECR) module (Sec.~\ref{sec:3.1}) to decompose physical phenomena into a sequence of fine-grained event units. To mitigate causal ambiguities, we embed computational analysis driven by physical formulas into the reasoning process with scene graphs. This enables the inference of physically realistic events with clear causal relationships. (2) We develop a Transition-aware Cross-modal Prompting (TCP) module (Sec.~\ref{sec:3.2}) to ensure causal coherence and visual continuity between generated events, through the synergy of semantic and visual prompts. From a semantic perspective, this module compresses multiple event descriptions into a single causally consistent representation using causal conjunctions. On the visual side, this module uses keyframes synthesized by interactive editing as visual prompts, maintaining the smooth transition between events. 

We evaluate our framework on PhyGenBench \cite{meng2024towards} and VideoPhy \cite{bansal2024videophy} benchmarks. Our framework significantly outperforms current PPVG approaches on physics-informed metrics across diverse physical domains. Crucially, videos generated by our framework can preserve reasonable chronological order of physical events. Our contributions are summarized as follows:
\begin{itemize}
\item We propose an event-centric generation framework that models physically plausible videos as sequences of causally connected and dynamically evolving events.
\item To address causal ambiguity, we decompose physical phenomena into causally ordered event units by causal reasoning with deterministic physical constraints.
\item To constrain continuous generation between physical events, we synthesize temporally aligned semantic-visual prompts to guide event transitions. 
\item Comprehensive experiments demonstrate that our framework outperforms existing methods in generating physically realistic and causally coherent videos.
\end{itemize}

 \section{Related Works}
\label{sec:related_works}
\textbf{Physical Plausible Video Generation.} To make videos obey physical laws, physics-aware generation has been explored. Several works \cite{hu2019difftaichi,hu2019taichi,hsu2025autovfx} characterize physical phenomena through simulations based on \textit{graphics engines}, which are integrated into diffusion sampling to enhance physical realism. To handle diverse \textit{open-domain physical phenomena}, VideoREPA \cite{zhang2025videorepa} leverages physical knowledge from foundation models. WISA \cite{wang2025wisa} and PhysHPO \cite{chen2025hierarchical} guide diffusion models to learn physical phenomena from decomposed principles. VLIPP \cite{yang2025vlipp}, DiffPhy \cite{zhang2025think}, PAG-SAD \cite{hao2025enhancing}, and PhyT2V \cite{xue2025phyt2v} leverage CoT reasoning to design physics-aware prompts. However, physical events unfold as causally ordered processes, while current methods, hindered by lack of causal modeling, often collapse them into a single scene.

\noindent \textbf{Chain-of-Thought in Visual Generation.} Recent studies have adapted CoT reasoning \cite{wei2022chain} from language understanding to visual generation, which are divided into two categories. The first leverages \textit{reasoning before generation} paradigm to augment conditioning signals \cite{huang2025interleaving,zeng2025draw,zhang2025layercraft,fang2025got}. For example, LayerCraft \cite{zhang2025layercraft}, and GoT \cite{fang2025got} enable the generation of multiple objects by reasoning about spatial arrangements. Others embed step-by-step reasoning into the synthesis process through \textit{reasoning during generation} paradigm. Z-Sampling \cite{bai2024zigzag} performs diffusion self-reflection, bridging the gap between denoising and inversion. Visual-CoG \cite{li2025visual} adopts a chain-of-guidance framework to supervise each generation stage. However, current approaches mainly focus on semantic and spatial reasoning, neglecting the modeling of deterministic causal relationships.

\noindent \textbf{Dual-Prompt in Video Generation.} While natural language defines scene semantic, it often underspecifies detailed geometry and motion. Accordingly, visual cues are introduced to guide video generation, including reference image, spatial layouts, and motion priors. Some works \cite{girdhar2024factorizing,lin2025stiv,lai2025incorporating,ni2024ti2v,li2023videogen} employ \textit{reference image} as the appearance prior for generating high-fidelity textures and diverse visual styles. To enhance geometric details, several studies introduce \textit{spatial layouts} during generation. SketchVideo \cite{liu2025sketchvideo} leverages sketches to constrain the contours of objects. DyST-XL \cite{he2025dyst} and BlobGEN-Vid \cite{feng2025blobgen} specify the locations of objects through bounding boxes and blobs, respectively. Given the dynamic nature of videos, recent studies use \textit{motion priors} to capture sophisticated trajectories \cite{zhou2025trackgo,he2024mojito}. However, these approaches primarily constrain individual scenes, lacking the ability to ensure smooth transitions between multiple events.

 \section{Methodology}
\subsection{Overall Framework}
Given a user-provided linguistic description $w$ of physical phenomenon, our goal is to generate the corresponding physically plausible video $\mathbf{V}$ which characterizes underlying progression of described phenomenon.
\begin{equation}
\Gamma: w \rightarrow \mathbf {V},    
\end{equation}
where $\Gamma$ denotes our physics-aware video generation framework. Specifically, our framework is organized as two synergistic modules. In ~\cref{sec:3.1}, we design a Physics-driven Event Chain Reasoning (PECR) module, which interprets each complex phenomenon described in user-provided description into an ordered collection of physical events. In ~\cref{sec:3.2}, we develop a Transition-aware Cross-modal Prompting (TCP) module, which bridges the event chain inferred by PECR module to the video generation process. Instead of time-invariant linguistic descriptions and reference images, our TCP module dynamically synthesizes dual-conditions evolving with physical processes. %Overall, our framework captures the continuous evolution of physical processes, allowing the generated videos to faithfully reproduce dynamic physical phenomena. 

\subsection{Physics-driven Event Chain Reasoning}
\label{sec:3.1}
\begin{figure*}[htbp]
\centering
  \includegraphics[width=\linewidth]{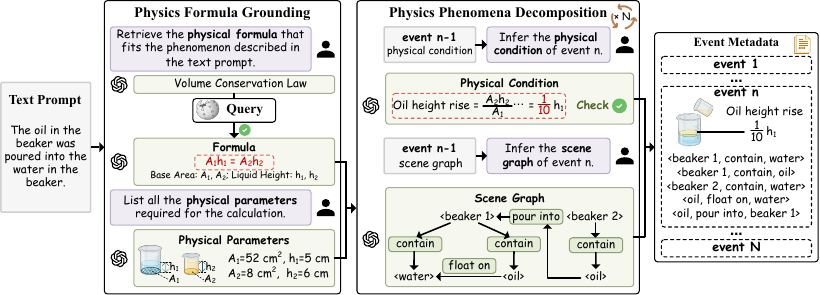}
  % \includesvg[width=\linewidth]{m1.svg}
  \caption{Overview of our PECR module (~\cref{sec:3.1}). This module conceptualizes physical phenomena in user-provided descriptions as a series of causally ordered events governed by real-world physical formulas, where each event encompasses semantic descriptions and measurable physical parameters for key objects. This characterizes the underlying scene changes induced by such phenomena.}
  \label{fig:module1}
\end{figure*}
Physical phenomenon involves the progression of events together with the corresponding changes of the physical parameters. Current studies \cite{zhang2025think,xue2025phyt2v} typically bind a physical phenomenon to an individual object, and simply use a semantic tag to coarsely describe each phenomenon. Unlike such approaches, we conceptualize physical phenomena as a series of causally ordered events, as shown in ~\cref{fig:module1}. Each event can be regarded as a composite unit, encompassing descriptive information of objects’ semantic and interactions, as well as measurable physical conditions governed by physics formulas.  This enables the characterization of crucial moments in a physical process from both qualitative and quantitative perspectives.

\noindent \textbf{Physics Formula Grounding.} To numerically describe physical processes, we perform reasoning over the physics formulas $\mathcal{F}^*$ based on the physical laws $\mathcal{L}$ embedded in the linguistic description $w$. Specifically, physical laws $\mathcal{L}$ are determined by question answering, where options are defined according to \cite{lin2025exploring}. Formula names associated with physical laws are inferred from the linguistic description. After that, inferred formula names $\mathcal{N}_{\mathcal{L}}$ are used as queries to retrieve physical formulas $\mathcal{F}^*$ from knowledge bases.
\begin{equation}
\mathcal{F}^* = \operatorname{TopK}_{f \in \mathcal{F}_{\mathcal{L}}} P(f \mid \mathcal{N}_{\mathcal{L}}, \mathcal{L}),
\end{equation}
where $\mathcal{F}_{\mathcal{L}}$ denotes all formulas in the online knowledge base associated with physical laws $\mathcal{L}$. $P(\cdot)$ is the scoring function over the candidate formulas $\mathcal{F}_{\mathcal{L}}$. When no direct match of inferred formula names $\mathcal{N}_{\mathcal{L}}$ is found in $\mathcal{F}_{\mathcal{L}}$, we regenerate formula names using $\mathcal{F}_{\mathcal{L}}$. Once the formula is retrieved, the physical parameters required for formula analysis are set by commonesense reasoning.

\noindent \textbf{Physical Phenomena Decomposition.} To characterize the scene changes induced by complex physical phenomena, we decompose these phenomena into an ordered sequence of key events $ \{\mathcal{E}_t\}_{t=1}^{T} =\{\{\mathcal{C}_t\}_{t=1}^{T}, \{\mathcal{G}_t\}_{t=1}^{T} \} $. Where $\{\mathcal{E}_t\}_{t=1}^{T}$ denotes the metadata of events, $\{\mathcal{C}_t\}_{t=1}^{T}$ specifies the physical conditions, $\{\mathcal{G}_t\}_{t=1}^{T}$ denotes the dynamic scene graph, and $T$ is the event number. 

Physical conditions are calculated on the basis of our retrieved physics formulas. Intermediate quantities produced during the analytical calculation provide additional physically meaningful signals. By analyzing whether significant variations occur in physical parameters, the boundaries of physical events can be determined. These boundaries enable the continuous video to be discretized into a sequence of events.
\begin{equation}
\mathcal{C}_t = \left\{ \big(\mathbf{P}_t, \mathcal{F}^*(\mathbf{P}_t)\big) \,\big|\, 
\|\mathbf{P}_t - \mathbf{P}_{t-1}\| > \tau_p \right\},
\end{equation}
where $\mathbf{P}_t$ denotes the physical parameter vectors of all objects within the $t$-th event. $\tau_p$ is the variation threshold, determining whether the change in physical parameters is sufficient to indicate a new event. In order to ensure physical consistency, the parameters inferred for current event are validated against those of neighboring events by detecting abrupt changes that violate physical continuity. When invalid changes are found, the corresponding parameters and physical contexts are fed back for re-inference. 

After that, we update the scene graph based on physical conditions. Given $ \mathcal{C}_t $, we update the nodes $ \mathcal{V}_t $ and edges $ \mathcal{R}_t $ of the scene graph $\mathcal{G}_t = \{\mathcal{V}_t, \mathcal{R}_t\}$. For nodes $ \mathcal{V}_t $, appearance (\textit{e.g.}, liquid changes color) or semantic label (\textit{e.g.}, burn to ashes) would be updated according to variations in their physical parameters. Update of edges $ \mathcal{R}_t $ is driven by changes of interactions between objects (\textit{e.g.}, a decrease in distance), which requires considering coordinated variations of physical parameters across multiple objects:
\begin{equation}
\mathcal{G}_t = \Phi(\mathcal{G}_{t-1}, \mathcal{C}_t),
\end{equation}
where $\Phi(\cdot)$ denotes the scene graph update function, which incorporates physical variations captured in $\mathcal{C}_t$ into previous scene graph $\mathcal{G}_{t-1}$ to produce current new graph $\mathcal{G}_t$.

\subsection{Transition-aware Cross-modal Prompting}
\label{sec:3.2}
\begin{figure*}[htbp]
\centering
  \includegraphics[width=\linewidth]{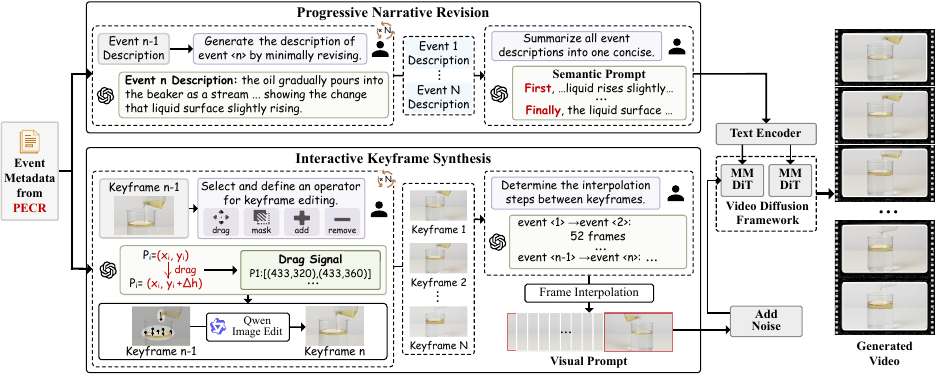}
  \caption{Overview of our TCP module (~\cref{sec:3.2}). This module aims to generate semantic-visual prompts for each event based on its meta data. Semantic prompts are inferred by our proposed progressive narrative revision, serving as the guidance during denoising steps. Visual prompts are obtained by our proposed interactive keyframe synthesis, replacing original noise to provide physics-aware priors.}
  \label{fig:module2}
\end{figure*}

With a sequence of events inferred from the previous stage, the key challenge is to bridge such events to physically realistic video generation. A dual-condition framework \cite{lin2025stiv,ni2024ti2v} serves as a promising solution. However, current frameworks fail to make conditions evolve over time, leading to an inability to capture events’ progression. Unlike prior work, we progressively synthesize semantic–visual prompts for each event while preserving seamless temporal progression, as shown in ~\cref{fig:module2}. The semantic prompt serves as a guidance during the denoising and the visual prompt replaces original Gaussian noise to provide physics-aware priors. Generation process is formally defined as:
\begin{equation}
\mathbf{Z}_{\tau_{z}-1} = \epsilon_\theta(\mathbf{Z}_{\tau_{z}}; \mathbf{W}),
\end{equation}
where $\mathbf{Z}_{\tau_{z}}$ denotes the visual priors. $\mathbf{W}$ specifies the embedding of linguistic description. $\epsilon_\theta $ is denoising network of the video diffusion model.

\noindent \textbf{Progressive Narrative Revision.} Describing events independently can disrupt holistic coherence of narratives. Therefore, we perform minimal progressive revisions conditioned on the preceding context. Specifically, physical conditions $\mathcal{C}_t$ constrain physically permissible transitions, \textit{e.g.}, rising temperature allows “melting” and excludes “freezing”. We employ scene graphs $\mathcal{G}_t$ to preserve object identities and to specify which attributes and relations may change. Let $w_t$ denotes the description of the $t$-th event, we obtain $w_t$ via:
\begin{equation}
w_t = \text{LLM}\big(w_{t-1} + \Delta(w_{t-1}, \mathcal{C}_t, \mathcal{G}_t)\big),
\end{equation}
where $\Delta(\cdot)$ denotes an incremental semantic revision guided by $\mathcal{C}_t$ and $\mathcal{G}_t$. 

Video diffusion models typically condition on a single sentence. However, concatenating multiple event descriptions produces semantic redundancy. Given this, we merge event descriptions into a positive semantic prompt via semantic condensation and causal connectives. Additionally, we construct a negative description $w^{*}_{-}$. Two descriptions are embedded separately and concatenated as:
\begin{equation}
\mathbf{W} = [\,\psi_{\text{text}}(w^{*}_{+});\; \psi_{\text{text}}(w^{*}_{-})\,],
\end{equation}
where $\psi_{\text{text}}(\cdot)$ is the text encoder.

\noindent \textbf{Interactive Keyframe Synthesis.} While language provides conceptual semantic guidance, physically realistic details remain under-specified, due to the ambiguity of linguistic description. To embed physics-aware details into random Gaussian noise, we synthesize keyframes $v_t$ for each physical event by interactive image editing. These generated keyframes serve as priors to guide video generation process. To be specific, we select the appropriate editing operator from a predefined set (\textit{e.g.}, drag or mask). The change in physical parameters across consecutive conditions acts as a numerical regularizer. It bounds the action space, including dragging magnitude and area of visual change. Practically, we draw operator cues on a source image to lead the modifications through Qwen-Image-Edit \cite{wu2025qwen}. Each keyframe is obtained as follows:
\begin{equation}
\mathcal{O}_t = \text{LLM}\big((\mathcal{C}_{t-1}, \mathcal{G}_{t-1}) \rightarrow (\mathcal{C}_t, \mathcal{G}_t)\big),
\end{equation}
\begin{equation}
v_t = \text{Edit}(v_{t-1}; \mathcal{O}_t)
\end{equation}
where $\mathcal{O}_t $ denotes the editing operator. The keyframe $ v_1$ is directly synthesized from the event description. 

To provide a smooth progression, we apply linear interpolation between keyframes. In parallel with inferring $\mathcal{O}_t$, a physically plausible time span $d_t$ for the variation from the $(t-1)$-th to the $t$-th event is also predicted, which determines how many in-between frames are interpolated between $ v_{t-1}$ and $v_t$. Because video diffusion models often operate in a compressed feature space, we use VAE \cite{kingma2013auto} to encode each keyframe. Based on predicted time span and embedded keyframe features, we perform frame interpolation as follows:
\begin{equation}
\mathbf{z}_{0,t} = \mathrm{INTERP}(\psi_{\text{img}}(v_{t-1}),\; \psi_{\text{img}}(v_t);\; d_t\big),
\end{equation}
where $\mathbf{z}_{0,t}$ denotes the interpolated features for variation between two events. $\mathrm{INTERP}(\cdot)$ specifies linear interpolation. $\psi_{\text{img}}(\cdot)$ is VAE encoder. Given a predefined timestep $\tau_{z}$, we add noise to $[\mathbf{z}_0,…, \mathbf{z}_T]$, serving as priors to denoising process of the video diffusion model.
\begin{equation}
\mathbf{Z}_{\tau_{z}} = [\mathbf{z}_0,…, \mathbf{z}_T] + \sigma_{\tau_{z}}^2 \epsilon, \quad \epsilon \sim \mathcal{N}(0, I),
\end{equation}
where $\sigma_{\tau_{z}}^2$ denotes the variance of Gaussian noise.
 \section{Experiments}
\subsection{Experimental Setups}
% This section sequentially describes used datasets and evaluation metrics in our experiments, and implementation details of our approaches.

\begin{figure*}[htbp]
  \centering
  % \includesvg[width=\textwidth]{result_1.svg}
  \includegraphics[width=1.0\textwidth]{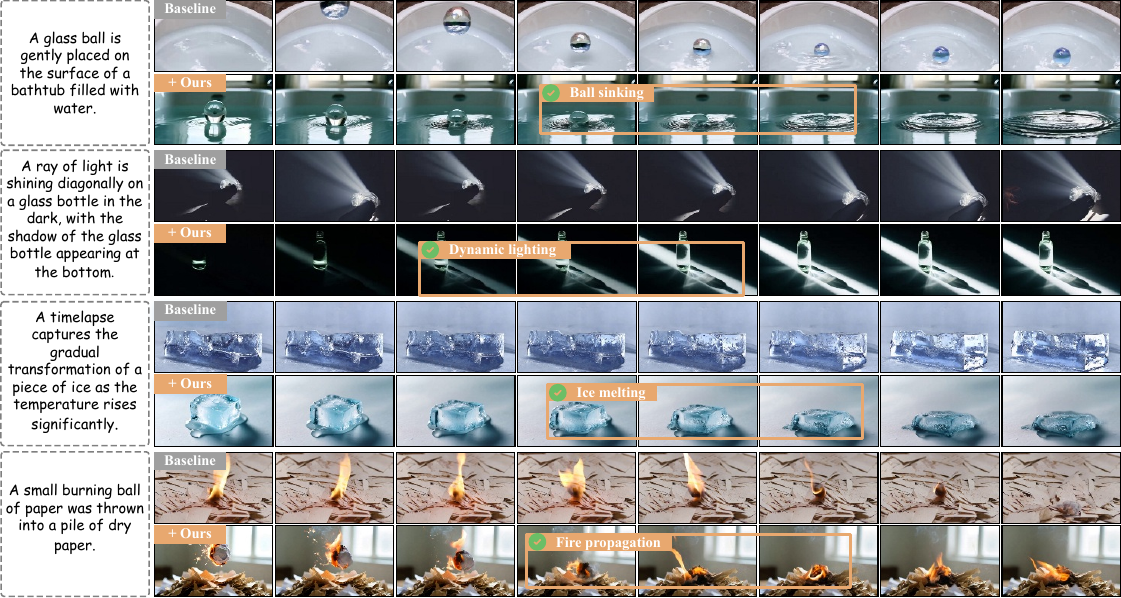}
  \caption{Visualization of physics-aware video generation results across four physical domains. Compared with baseline CogVideo-5B \cite{yang2024cogvideox}, our approach yields causally coherent progressions of physical phenomena, \textit{e.g.}, the glass ball sinks, the bottom shadow extends in the direction of the light, gradual melting of ice, fire spreading through the paper. All prompts are sourced from PhyGenBench \cite{meng2024towards}. }
  \label{fig:vis_phygenbench}
\end{figure*}

\noindent \textbf{Datasets.} We evaluate on PhyGenBench \cite{meng2024towards} and VideoPhy \cite{bansal2024videophy} datasets. Specifically, PhyGenBench comprises 160 designed linguistic descriptions spanning 27 physical laws across four fundamental domains, namely mechanics, optics, thermal, and material. VideoPhy provides a collection of 688 human-verified linguistic prompts, describing various physical interactions between objects, comprising solid-solid, solid-fluid, and fluid-fluid.

\noindent \textbf{Evaluation Metrics.} Following PhyGenBench \cite{meng2024towards}, we use Physical Commonsense Alignment (PCA) as our metric, which indicates video quality by considering key phenomena detection, physics order verification, and overall naturalness evaluation. For VideoPhy \cite{bansal2024videophy}, we use its provided VideoCon-Physics evaluator to assess Semantic Adherence (SA) and Physical Commonsense (PC). SA evaluates whether a linguistic description is semantically grounded in generated video frames. PC examines whether the depicted actions and object properties conform to real-world physics laws.

\noindent \textbf{Implementation Details.} We use CogVideoX 5B \cite{yang2024cogvideox} as our video generation baseline. Following the official implementation, the video generation model is configured with 161 frames per sample and a resolution of 1360$\times$768. For language reasoning, we employ an open-source GPT-OSS-20B \cite{openai2025gptoss20b} with the default configuration. We use the Qwen-Image family \cite{wu2025qwen} for keyframe generation. 
% All experiments are run on a single NVIDIA H100GPUs (80G). 

\begin{table}[htbp]
\centering
\caption{Performance comparison on PhyGenBench \cite{meng2024towards} across four physical domains. The best and second-best results are \textbf{highlighted} and \underline{underlined}, respectively.}
\small
\resizebox{\columnwidth}{!}{%
\begin{tabular}{lccccc}
\toprule
\multirow{2}{*}{Methods} & \multicolumn{4}{c}{Physical domains ($\uparrow$)} & \multirow{2}{*}{Avg.} \\
\cmidrule{2-5}
 & Mechanics & Optics & Thermal & Material &  \\
\midrule
\multicolumn{6}{l}{\textit{Video Foundation Model}} \\
\textcolor{black}{Lavie \cite{wang2025lavie}} & \textcolor{black}{0.30} & \textcolor{black}{0.44} & \textcolor{black}{0.38}
& \textcolor{black}{0.32} & \textcolor{black}{0.36} \\

% \textcolor{black}{CogVideoX-2B \cite{yang2024cogvideox}} & \textcolor{black}{0.38} & \textcolor{black}{0.43} & \textcolor{black}{0.34}
% & \textcolor{black}{0.39} & \textcolor{black}{0.37} \\

\textcolor{black}{VideoCrafter v2.0 \cite{chen2024videocrafter2}} & - & - & - & - & \textcolor{black}{0.48} \\

\textcolor{black}{Open-Sora v1.2 \cite{zheng2024open}} & \textcolor{black}{0.43} & \textcolor{black}{0.50} & \textcolor{black}{0.34}
& \textcolor{black}{0.37} & \textcolor{black}{0.44} \\

\textcolor{black}{Vchitect v2.0 \cite{fan2025vchitect}} & \textcolor{black}{0.41} & \textcolor{black}{0.56} & \textcolor{black}{0.44}
& \textcolor{black}{0.37} & \textcolor{black}{0.45} \\

\textcolor{black}{Wan \cite{wan2025wan}} & \textcolor{black}{0.36} & \textcolor{black}{0.53} & \textcolor{black}{0.36}
& \textcolor{black}{0.33} & \textcolor{black}{0.40} \\

\textcolor{black}{Kling \cite{klingai2024}} & \textcolor{black}{0.45} & \textcolor{black}{0.58} & \textcolor{black}{0.50}
& \textcolor{black}{0.40} & \textcolor{black}{0.49} \\

\textcolor{black}{Pika \cite{Pika2024}} & \textcolor{black}{0.35} & \textcolor{black}{0.56} & \textcolor{black}{0.43}
& \textcolor{black}{0.39} & \textcolor{black}{0.44} \\

\textcolor{black}{Gen-3 \cite{runway2024gen3alpha}} & \textcolor{black}{0.45} & \textcolor{black}{0.57} & \textcolor{black}{0.49}
& \textcolor{black}{0.51} & \textcolor{black}{0.51} \\

\midrule
\multicolumn{6}{l}{\textit{Physics-aware Video Generation Model}} \\
% Cosmos-Diffusion-7B \cite{agarwal2025cosmos} & - & - & - & - & 0.24 \\
WISA \cite{wang2025wisa}                         & -    & -    & -    & -    & 0.43 \\
DiffPhy \cite{zhang2025think}                    & 0.53 & 0.59 & \underline{0.58} & 0.46 & 0.54 \\ 
CogVideoX-5B \cite{yang2024cogvideox}            & 0.39 & 0.55 & 0.40 & 0.42 & 0.45 \\
\, + PhyT2V \cite{xue2025phyt2v}                 & 0.45 & 0.55 & 0.43 & 0.53 & 0.50 \\
\, + SGD \cite{hao2025enhancing}                 & 0.49 & 0.58 & 0.42 & 0.48 & 0.49 \\
%\, + Vanilla DPO \cite{Wallace2024DiffusionDPO}  & 0.48 & 0.60 & 0.47 & 0.58 & 0.54 \\
\, + PhysHPO \cite{chen2025hierarchical}         & \underline{0.55} & \underline{0.68} & 0.50 & \textbf{0.65} & \underline{0.61} \\
\rowcolor{RowHL}
\, + Ours          & \textbf{0.67} & \textbf{0.72} & \textbf{0.65} & \underline{0.60} & \textbf{0.66} \\
\bottomrule
\end{tabular}%
}
\label{tab:phygenbench}
\end{table}

\begin{table*}[htbp]
\centering
\caption{Performance comparisons on VideoPhy \cite{bansal2024videophy} across various physical interactions between objects. The best and second-best results are \textbf{highlighted} and \underline{underlined}, respectively.}
\small
\begin{tabularx}{\textwidth}{l*{12}{Y}}
\toprule
\multirow{2}{*}{Methods} & \multicolumn{3}{c}{Overall (\%)} & \multicolumn{3}{c}{Solid-Solid (\%)} & \multicolumn{3}{c}{Solid-Fluid (\%)} & \multicolumn{3}{c}{Fluid-Fluid (\%)} \\
\cmidrule(lr){2-4}\cmidrule(lr){5-7}\cmidrule(lr){8-10}\cmidrule(lr){11-13}
& \SAPC & SA & PC & \SAPC & SA & PC & \SAPC & SA & PC & \SAPC & SA & PC \\
\midrule
\multicolumn{13}{l}{\textit{Video Foundation Model}} \\
\grayrowA{VideoCrafter2 \cite{chen2024videocrafter2}}{19.0}{48.5}{34.6}{4.9}{31.5}{23.8}
\grayrowB{27.4}{57.5}{41.8}{32.7}{69.1}{43.6}

\grayrowA{LaVIE \cite{wang2025lavie}}{15.7}{48.7}{28.0}{8.5}{37.3}{19.0}
\grayrowB{15.8}{52.1}{30.8}{34.5}{69.1}{43.6}

\grayrowA{SVD-T2I2V \cite{blattmann2023stable}}{11.9}{42.4}{30.8}{4.2}{25.9}{27.3}
\grayrowB{17.1}{52.7}{32.9}{18.2}{58.2}{34.5}

%\grayrowA{ZeroScope \cite{cerspenseZeroscope576w2023}}{11.9}{30.2}{32.6}{6.3}{17.5}{22.4}
%\grayrowB{14.4}{40.4}{37.0}{20.0}{36.4}{47.3}

\grayrowA{OpenSora \cite{zheng2024open}}{4.9}{18.0}{23.5}{1.4}{7.7}{23.8}
\grayrowB{7.5}{30.1}{21.9}{7.3}{12.7}{27.3}

\grayrowA{Pika \cite{Pika2024}}{19.7}{41.1}{36.5}{13.6}{24.8}{36.8}
\grayrowB{16.3}{46.5}{27.9}{44.0}{68.0}{58.0}

\grayrowA{Dream Machine \cite{LumaDreamMachine2024}}{13.6}{61.9}{21.8}{12.6}{50.0}{24.3}
\grayrowB{16.6}{68.1}{23.6}{9.0}{76.3}{11.0}

%\grayrowA{Lumiere-T2I2V \cite{bar2024lumiere}}{12.5}{45.8}{25.0}{8.1}{37.1}{25.2}
%\grayrowB{17.1}{59.6}{26.0}{10.9}{49.1}{21.8}

\grayrowA{Lumiere \cite{bar2024lumiere}}{9.0}{38.4}{27.9}{8.4}{26.6}{27.3}
\grayrowB{9.6}{47.3}{26.0}{9.1}{45.5}{34.5}

%\grayrowA{Gen-2 \cite{esser2023structure}}{7.6}{26.6}{27.2}{4.0}{8.9}{37.1}
%\grayrowB{8.1}{38.5}{18.5}{15.1}{37.7}{26.4}
\midrule
\multicolumn{13}{l}{\textit{Physics-aware Video Generation Model}} \\
CogVideoX-5B \cite{yang2024cogvideox}            & 39.6 & 63.3 & 53 & 24.4 & 50.3 & 43.3 & 53.1 & 76.5 & 59.3 & 43.6 & 61.8 & \textbf{61.8} \\
\, + PhyT2V \cite{xue2025phyt2v}                 & 40.1 & - & - & 25.4 & - & - & 48.6 & - & - & 55.4 & - & - \\
\, + Vanilla DPO \cite{Wallace2024DiffusionDPO}  & 41.3 & - & - & 28.2 & - & - & 50.0 & - & - & 51.8 & - & - \\
% \, + PhysHPO \cite{chen2025hierarchical}      & \underline{45.9} & - & - & \underline{32.4} & - & - & \underline{54.1} & - & - & \underline{58.9} & - & - \\
\rowcolor{RowHL}
\, + Ours         & \textbf{49.3} & \textbf{79.5} & \textbf{59.4} & \textbf{40.6} & \textbf{73.4} & \textbf{53.8} & \textbf{60.0} & \textbf{85.6} & \textbf{66.7} & \underline{54.5} & \textbf{85.4} & \textbf{61.8} \\
\bottomrule
\end{tabularx}
\label{tab:videophy}
\end{table*}
\begin{figure*}[htbp]
  \centering
  % \includesvg[width=\textwidth]{result_2.svg}
  \includegraphics[width=1.0\textwidth]{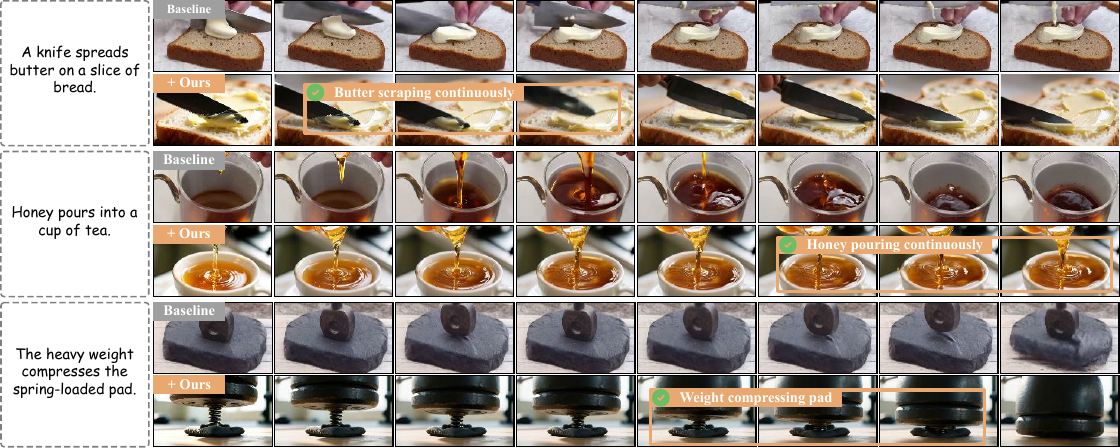}
  \caption{Visualization of physics-aware video generation results across various physical interactions between objects. Compared with baseline CogVideo-5B \cite{yang2024cogvideox}, our approach demonstrates clearer causal progression, \textit{e.g.}, butter spread along the knife movement, continuous honey inflow with a rising level, and monotonic spring compression. All prompts are sourced from VideoPhy \cite{bansal2024videophy}.}
  \label{fig:vis_videophy}
\end{figure*}

\subsection{Evaluation on PhyGenBench}
We compare our framework with video foundation models \cite{wang2025lavie,chen2024videocrafter2,zheng2024open,fan2025vchitect,wan2025wan,klingai2024,runway2024gen3alpha,yang2024cogvideox} and physics-aware video generation models \cite{wang2025wisa,zhang2025think,xue2025phyt2v,hao2025enhancing,chen2025hierarchical} on the PhyGenBench \cite{meng2024towards} benchmark. As shown in ~\cref{tab:phygenbench}, our framework consistently achieves the best overall performance of 0.66, surpassing PhysHPO \cite{chen2025hierarchical} (previous SOTA) by 8.19\% on average. As shown in ~\cref{fig:vis_phygenbench}, compared with the baseline CogVideo-5B \cite{yang2024cogvideox}, our framework achieves visually more realistic generation of gradual sinking (row 1), light refraction (row 2), realistic melting (row 3), and natural combustion (row 4). These results indicate the ability of our framework to understand physical laws and their corresponding visual dynamics by explicitly considering causally ordered events.
% Detailed PhyGenBench evaluations of key phenomena and physical orders are provided in ~\cref{tab:phygenbench_additional}.
\subsection{Evaluation on VideoPhy}
We compare our framework with some video generation models \cite{chen2024videocrafter2,wang2025lavie,blattmann2023stable,zheng2024open,Pika2024,LumaDreamMachine2024,bar2024lumiere,yang2024cogvideox,chen2025hierarchical,xue2025phyt2v} on the VideoPhy \cite{bansal2024videophy} benchmark. As shown in ~\cref{tab:videophy}, our approach achieves 49.3\% scores (SA=1, PC=1) in general, significantly outperforming previous SOTA approach PhysHPO \cite{chen2025hierarchical} by approximately 3.4\%. This confirms the effectiveness of our framework in adhering to the semantic cues and following physical laws by leveraging vision-language prompts inferred via CoT reasoning. As shown in ~\cref{fig:vis_videophy}, compared with the baseline CogVideo-5B \cite{yang2024cogvideox}, our approach shows butter spreads along the knife path (row 1), honey pouring as a continuous viscous stream with a steadily rising liquid level (row 2), and the spring shortens monotonically under sustained compression (row 3). These cases emphasize that our framework can capture interactions between objects in diverse physical phases.

\subsection{Ablation Studies}
We perform ablation studies based on PhyGenBench \cite{meng2024towards} to systematically analyze the contributions of each block in our proposed PECR and TCP modules.
\begin{table}[htbp]
\centering
\caption{Ablation analysis of PECR and TCP modules, including Physics Formula Grounding (PFG) and Physical Phenomena Decomposition (PPD) in PECR, and Progressive Narrative Revision (PNR) and Interactive Keyframe Synthesis (IRS) in TCP.}
\label{tab:ablation_study}
\small
\resizebox{\columnwidth}{!}{%
\begin{tabular}{lccccc}
\toprule
\multirow{2}{*}{Variant} & \multicolumn{4}{c}{Physical domains ($\uparrow$)} & \multirow{2}{*}{Avg.} \\
\cmidrule{2-5}
 & Mechanics & Optics & Thermal & Material &  \\
\midrule
\rowcolor{RowHL}
Ours  & \textbf{0.67} & \textbf{0.72} & \textbf{0.65} & \textbf{0.60} & \textbf{0.66} \\
\midrule
% \addlinespace[2pt]
\multicolumn{6}{l}{\textit{Ablations of PECR module}} \\
w/o PFG & 0.63 & 0.69 & 0.61 & 0.53 & 0.62 \\
w/o PPD & 0.58 & 0.67 & 0.61 & 0.52 & 0.59 \\
\midrule
% \addlinespace[2pt]
\multicolumn{6}{l}{\textit{Ablations of TCP module}} \\
w/o PNR & 0.65  & 0.70 & 0.64  & 0.56  & 0.64 \\
w/o IKS & 0.50 & 0.64 & 0.58 & 0.48 & 0.55  \\
\bottomrule
\end{tabular}%
}
\end{table}
\begin{figure}[htbp]
  \centering
  \includegraphics[width=\columnwidth]{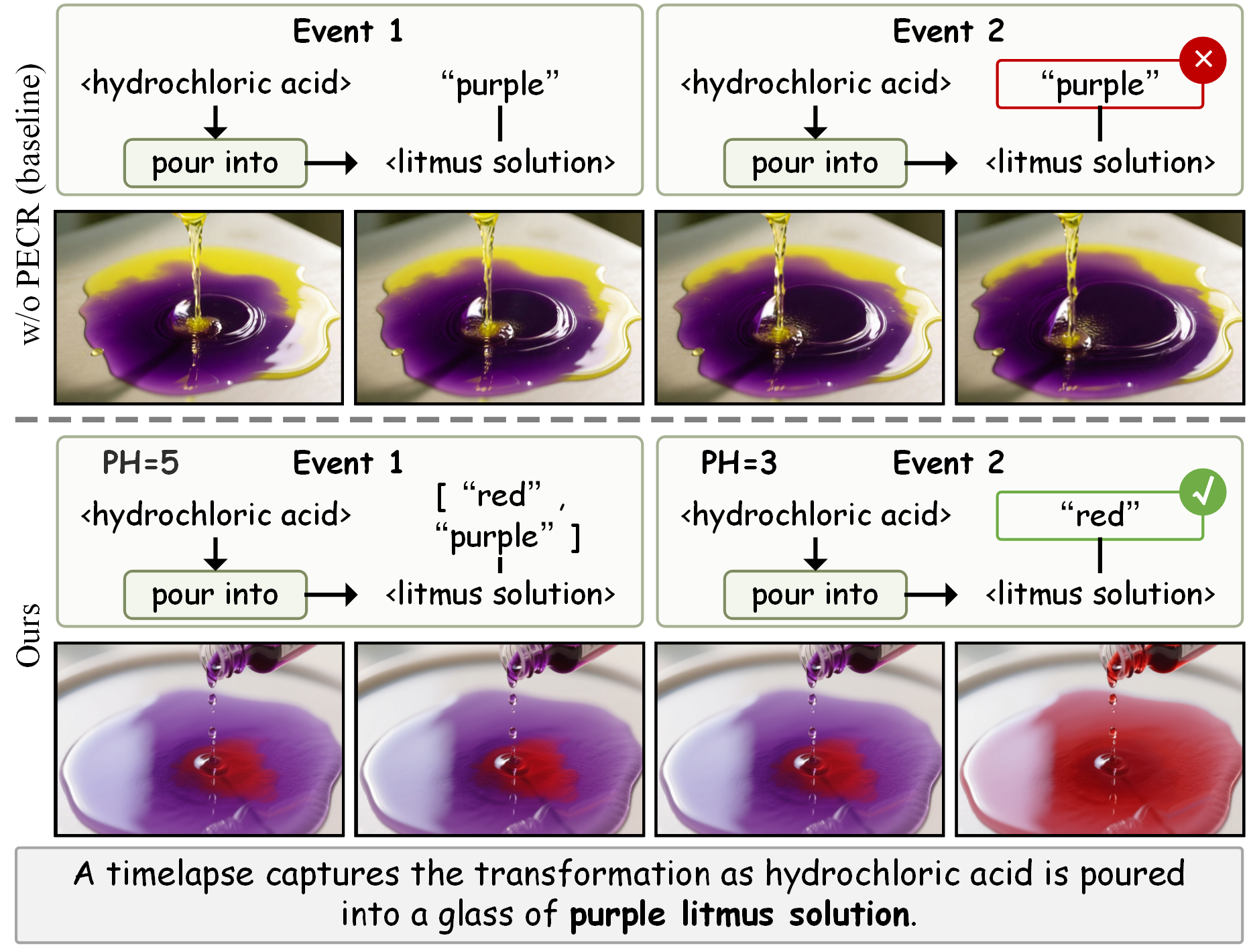}
  \caption{Ablation of the PECR module. Physical-driven CoT reasoning in PECR is crucial for modeling gradual changes in physical conditions and preserving causal consistency across events.}
  \label{fig:sm_ablation_1}
\end{figure}

\noindent \textbf{PFG and PPD blocks in PECR.} We analyze the effectiveness of the Physics Formula Grounding (PFG) and Physical Phenomena Decomposition (PPD) blocks in PECR module. When removing the PFG block, physical events are inferred solely from original descriptions; whereas without PPD block, cross-modal conditions are derived only from physical formulas. As shown in ~\cref{tab:ablation_study}, excluding the PFG block causes an average performance drops of about 6\% across physical domains, underscoring the necessity of formulas to quantitatively understand physical laws. Similarly, removing the PPD block leads to an average decline of around 11\%, indicating its effectiveness in generating realistic physical progressions by decomposing complex processes into logically ordered event chains. 

To validate the role of PECR module in our overall framework, we disable its CoT reasoning and verification procedures. In this setup, the module directly infers the scene graph of individual events from original descriptions. As shown in \cref{fig:sm_ablation_1}, our framework correctly generates the gradual color transition in the litmus solution from purple to red, whereas the baseline produces a persistently red solution. Upon inspecting inferred scene graphs, we observe that the PECR module updates the color property of the litmus node across events, but the variant leaves it unchanged. This indicates that PECR helps capture the causal progression of events by modeling how physical conditions evolve across events.

\begin{figure}[htbp]
  \centering
  \includegraphics[width=\columnwidth]{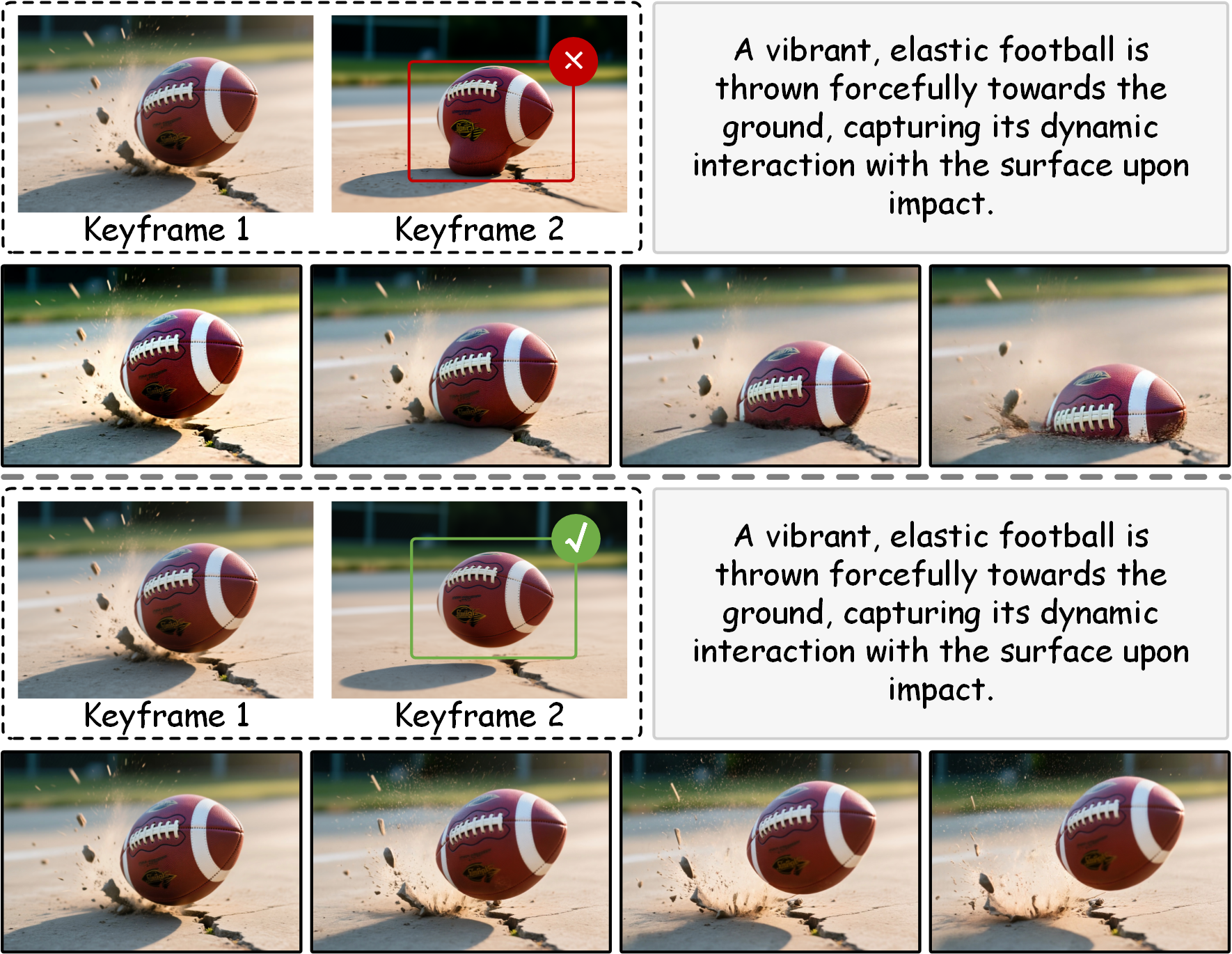}
  \caption{Ablation of visual prompts in TCP module. Physically consistent visual keyframe prompts are crucial, as unrealistic keyframes mislead the generator and lead to structurally implausible content.}
  \label{fig:sm_ablation_22}
\end{figure}

\noindent \textbf{PNR and IKS blocks in TCP.} We investigate the impact of the Progressive Narrative Revision (PNR) and Interactive Keyframe Synthesis (IRS) blocks in TCP module. As shown in ~\cref {tab:ablation_study}, without PNR block causes a moderate average drop of about 3\%, demonstrating its supportive role in improving the continuity and smooth evolution of scenes. Conversely, excluding the IKS block results in a significant average decrease of approximately 17\%, which underscores the essential role of explicitly generating dedicated keyframes for each physical phase in anchoring cross-frame dynamics and preserving a physically grounded visual progression. We also evaluate the effectiveness of visual prompts in TCP module. As shown in \cref{fig:sm_ablation_22}, the rounded shape of the rugby ball is preserved when keyframes provide reasonable visual cues. Conversely, the baseline produces an implausible result in which the ball sinks into the ground rather than resting on the surface. This result shows that the physical realism of visual prompts directly affects the structural fidelity of the generated content.

% \begin{figure}[htbp]
% \centering
%   % \includesvg[width=\linewidth]{event_curve.svg}
%   \includegraphics[width=1.0\linewidth]{figures/event_curve.pdf}
%   \caption{Effect of event number on Physical Commonsense Alignment (PCA) Score \cite{meng2024towards} across four physical domains: Mechanics, Optics, Thermal, and Material.}
%   \label{fig:event_curve}
% \end{figure}
% \noindent \textbf{Analysis of Event Number.} As shown in ~\cref {fig:event_curve}, our framework achieves the best performance with 4 events across all physical domains. Since each event serves as a guidance, too few events (e.g., 1–3) provide weak temporal supervision, making it difficult for the model to follow instructions accurately, thus reducing PCA scores. Meanwhile, increasing the number of events (e.g., 5–6) introduces accumulated errors in keyframe generation due to editing-based propagation, leading to poor leading signals and degraded video quality. To balance temporal guidance and keyframe stability, we set the event number to 4.

 \section{Conclusion}
This paper addresses the challenge of generating physics-aware videos characterizing the ordered progression of physical phenomena governed by real-world physical laws. In place of depicting a complex physical phenomenon by simple semantic labels, we decompose each phenomenon into a sequence of causally linked physical events based on standard physical formulas. We infer physical events and translate them into vision-language prompts that correspond to each event. These prompts are then used to condition the video generation process, ensuring alignment with physical dynamics. Comprehensive experiments confirm the effectiveness of our framework in generating physically plausible videos, especially in modeling complex and evolving physical phenomena. 
% However, as shown in ~\cref{fig:failure_case}, our framework occasionally fails in scenarios governed by compositional physical laws, as off-the-shelf foundation models have a weak capability in compositional physical reasoning \cite{qiu2025phybench,shi2025cryptox,zhang2025morpheus}. In the future, we plan to leverage advances in compositional visual reasoning \cite{ke2025explain,wang2025training} to enhance multi-physics consistency in physics-aware video generation.
% Besides, we enable such inferred physical events to be authentically realized as video dynamics by progressively synthesizing vision-language prompts for individual events and conditioning video generation process accordingly.
% \begin{figure}[htbp]
%   \centering
%   \includegraphics[width=\columnwidth]{figures/failure_case.pdf}
%   \caption{Failure case. As foundation models lack capability in compositional physical reasoning, our framework fails to generate scenarios governed by multiple physical laws.}
%   \label{fig:failure_case}
% \end{figure}

\noindent \textbf{Acknowledgement:} This work was supported by the National Natural Science Foundation of China (No. U23B2013, 62276176). This work was also partly supported by the SICHUAN Provincial Natural Science Foundation (No. 2024NSFJQ0023).

\balance
\clearpage
{
    \small
    \bibliographystyle{unsrt}
    \bibliography{main}
}

\end{document}